\documentclass{article}

    \usepackage[preprint,nonatbib]{neurips_2020}

\usepackage[utf8]{inputenc} % allow utf-8 input
\usepackage[T1]{fontenc}    % use 8-bit T1 fonts
\usepackage{hyperref}       % hyperlinks
\usepackage{url}            % simple URL typesetting
\usepackage{booktabs}       % professional-quality tables
\usepackage{amsfonts}       % blackboard math symbols
\usepackage{nicefrac}       % compact symbols for 1/2, etc.
\usepackage{microtype}      % microtypography
\usepackage{diagbox}
\usepackage{multirow}
\usepackage{graphicx}
\usepackage{float}
\usepackage{enumitem}
\usepackage{wrapfig}

\title{Transfer Learning or Self-supervised Learning? A Tale of Two Pretraining Paradigms}

\author{%
 Xingyi Yang\thanks{Equal Contribution}  , Xuehai He$^*$,Yuxiao Liang, Yue Yang \\
  University of California San Diego \\
  \texttt{\{x3yang,x5he,yul154\}@eng.ucsd.edu},\texttt{yuey9923@gmail.com} \\
  \And
  Shanghang Zhang \\
  University of California Berkeley \\
  \texttt{shz@eecs.berkeley.edu} \\
      \And
  Pengtao Xie \\
  University of California San Diego \\
  \texttt{pengtaoxie2008@gmail.com} \\
}

\begin{document}

\maketitle

\begin{abstract}
 Pretraining has become a standard technique in computer vision and natural language processing, which usually helps to improve performance substantially. Previously, the most dominant pretraining method is transfer learning (TL), which uses labeled data to learn a good representation network. Recently, a new pretraining approach -- self-supervised learning (SSL) -- has demonstrated promising results on a wide range of applications. SSL does not require annotated labels. It is purely conducted on input data by solving auxiliary tasks defined on the input data examples. The current reported results show that in certain applications, SSL outperforms TL and the other way around in other applications. There has not been a clear understanding on what properties of data and tasks render one approach outperforms the other. Without an informed guideline, ML researchers have to try both methods to find out which one is better empirically. It is usually  time-consuming to do so. In this work, we aim to address this problem. We perform a comprehensive comparative study between SSL and TL regarding which one works better under different properties of data and tasks, including domain difference between source and target tasks, the amount of pretraining data, class imbalance in source data, and usage of target data for additional pretraining, etc. The insights distilled from our comparative studies can help ML researchers decide which method to use based on the properties of their applications. 
\end{abstract}

\section{Introduction}
\vspace{-0.3cm}
Pretraining is a commonly used technique in deep learning to learn more effective representations for alleviating overfitting. Given a target task where the amount of training data is limited, training deep neural networks on this small-sized dataset has high risk of overfitting. To address this problem, one can pretrain the feature extraction layers in the network on large-sized external data from some source tasks, then finetune these layers on the target data. The abundance of source data enables the network to learn powerful representations that are robust to overfitting. And such representation power can be leveraged to assist in the learning of the target task with more resilience to overfitting.

Arguably, the most popular pretraining approach is transfer learning (TL)~\cite{tl}, which learns the weight parameters of a representation network by solving a supervised source task, i.e., correctly mapping input data examples to their labels (e.g., classes, segmentation masks, etc.). While successful, one concern of TL is that it uses labels in the source task to learn network weights, which may be biased to the source labels and generalize less well on the target task where the classes in the target labels are different from those in the source task. 

This problem can be potentially alleviated by unsupervised pretraining, which trains the network weights purely based on input data examples without using any labels in the source task. Recently, self-supervised learning (SSL)~\cite{he2019momentum,chen2020simple}, as an unsupervised pretraining approach, has achieved promising success and outperforms transfer learning in a wide range of applications~\cite{he2019momentum}. Similar to TL, SSL also solves predictive tasks. But the output labels in SSL are constructed from the input data, rather than annotated by human as in TL. The auxiliary predictive tasks in SSL could be predicting whether two augmented data examples originate from the same original data example~\cite{chen2020simple}, inpainting masked regions in images~\cite{inpainting}, etc. Since SSL does not leverage labels provided by human, it does not have the risk of being biased to labels in a source task. On the other hand, the potential pitfall of not using human-annotated labels is that the learned representations by SSL may not be as discriminative as those in TL. In sum, conceptually, SSL and TL both have advantages and disadvantages. It is difficult to judge whether one is better than the other. The existing empirical results show that in certain tasks, SSL outperforms TL~\cite{he2019momentum}; in other tasks, TL performs better than SSL~\cite{doersch2017multi}. But these studies did not provide a clear guidance on what properties of data and tasks render one approach works better than the other. Consequently, ML researchers are not informed about how to select the right pretraining method and they have to try both empirically to find out, which is time-consuming and resource-intensive.

To address this issue, we perform comprehensive studies to compare SSL and TL, and investigate which method works better under different properties of data and tasks, including domain difference between source and target tasks, the amount of pretraining data, class imbalance in source tasks, and the usage of target data for additional pretraining.  The studies are performed on 5 source tasks and 4 target tasks from various domains including daily-life objects, general scenes, natural creatures, and medical imaging. We summarize the insights distilled from the comparative studies to help ML researchers decide which pretraining method to use based on the specific properties of their applications.  
%In practice, between supervised transfer learning and unsupervised self-supervised learning, which pretraining approach should we use? Under different scenarios of domain discrepancy, data size, class space, which approach is more appropriate? Can they be combined together to yield better representations? In this work, we aim to address these questions by having comprehensive comparative studies.
The major insights obtained from the studies include: 
\begin{itemize}[leftmargin=*]
\setlength\itemsep{0em}
    \item When the domain difference between the source task and the target task is large, SSL outperforms TL. When domain difference is small, TL outperforms SSL.
    \item On the same source task, when the amount of pretraining data is small, SSL outperforms TL. When the amount is large, TL outperforms SSL.
\item SSL is less sensitive to domain difference than TL: on the same target task, the performance of SSL pretrained on source tasks with varying domain difference with the target task is relatively stable while that of TL varies a lot.
\item When domain difference is small, SSL is less sensitive to the amount of pretraining data than TL: given a target task and a source task that have small domain difference, the performance of TL on the target task changes a lot under varying amount of pretraining data in the source task while SSL is relatively stable.
\item SSL is more robust to class imbalance compared with TL: pretrained on source datasets with varying ratios of class imbalance, the performance of SSL varies less than TL.
\item For SSL, using the training examples in the target task as additional pretraining data works better than using source data only, which is not the case for TL.
\end{itemize}

%The rest of the paper of organized as follows. Section 2 presents an introduction of TL and SSL. Section 3 gives the design of comparative study. Section 4 presents the results. Section 5 reviews related works and Section 6 concludes the paper. 

\vspace{-0.3cm}
\section{Transfer Learning and Self-supervised Learning}
\vspace{-0.3cm}

\begin{wrapfigure}{r}{0.75\textwidth}
	\begin{center}
		\vspace{-0.3cm}
 	\includegraphics[width = 0.75\columnwidth]{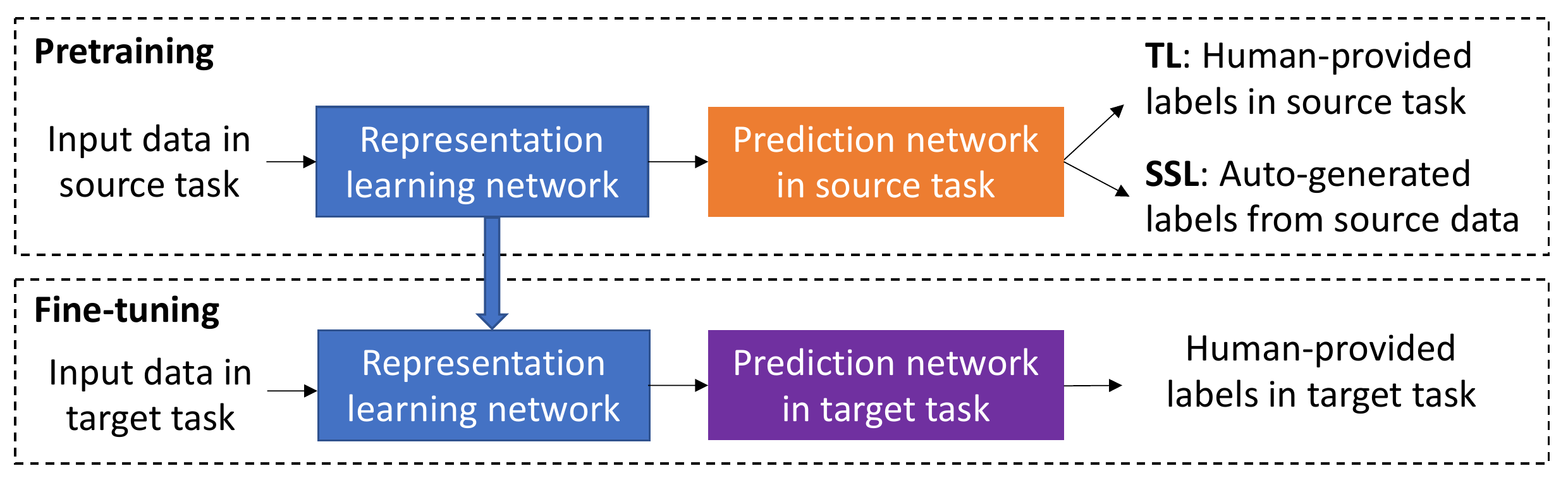}
 		\vspace{-0.4cm}
 	\caption{ Workflow of transfer learning and self-supervised learning. 
 	%It pretrains the weights of a representation network in a source task with labels. Then it finetunes the weights of this representation network in the target task.
	}\label{fig:tl}
	\end{center}
	\vspace{-0.4cm}
 \end{wrapfigure}
 
Figure~\ref{fig:tl} illustrates transfer learning (TL) and self-supervised learning (SSL). 
Both TL and SSL consist of two phrases: pretraining on source task and finetuning on target task. The purpose of pretraining is to train the weight parameters of a representation network into a good state. The pretrained weights are used to initialize the representation network in the target task (which has the same architecture as that in the source task). The initialized representation network in the target task is further trained on the target data. The difference between TL and SSL is that: TL performs pretraining using labeled source data in a supervised way while SSL performs pretraining using unlabeled source data in an unsupervised way.

Arguably, transfer learning~\cite{tl} is the most widely used method for pretraining. Given data examples $\mathcal{D}_T$ and their class labels $\mathcal{L}_T$ in a target task $T$, a neural network $\mathcal{N}_T$ is designed to fulfill this task. $\mathcal{N}_T$ is composed of two parts: a feature extraction sub-network $\mathcal{F}_T$ that learns latent feature representations of $\mathcal{D}_T$ and a prediction sub-network $\mathcal{P}_T$ which is tailored to the predictive task in $T$. Typically, the majority of weight parameters of $\mathcal{N}_T$ lie in the feature extraction part $\mathcal{F}_T$. Meanwhile, one has access to data examples $\mathcal{D}_S$ and their class  labels $\mathcal{L}_S$ in a source task $S$, where $\mathcal{D}_S$ is in general much larger than $\mathcal{D}_T$. How transfer learning (TL) works is: it creates a network $\mathcal{N}_S$ to solve the predictive task in $S$ where the feature extraction sub-network $\mathcal{F}_S$  has the same architecture as $\mathcal{F}_T$, but the prediction sub-network $\mathcal{P}_S$ is tailored to task $S$. TL trains $\mathcal{N}_S$ on $\mathcal{D}_S$ and $\mathcal{L}_S$ until convergence. Then it uses the weights of $\mathcal{F}_S$ to initialize $\mathcal{F}_T$ and trains $\mathcal{N}_T$ on $\mathcal{D}_T$ and $\mathcal{L}_T$ until convergence. While TL has demonstrated pervasive effectiveness, there is a major concern about this approach. The weights in $\mathcal{F}_S$ are trained by fitting the labels $\mathcal{L}_S$. Therefore  they are naturally biased to the classes in $\mathcal{L}_S$. When applied to predicting a new set of classes in $\mathcal{L}_T$, this bias may render the prediction less accurate. 
%Second, the source data $\mathcal{D}_S$ usually has a large domain discrepancy with $\mathcal{D}_T$. As a result, $\mathcal{P}_S$ trained on $\mathcal{D}_S$ may be biased to the data patterns in $\mathcal{D}_S$ and hence less capable of representing $\mathcal{D}_T$. 

 \iffalse
\begin{wrapfigure}{r}{0.8\textwidth}
	\begin{center}
	\vspace{-0.3cm}
 	\includegraphics[width = 0.8\columnwidth]{figs/ssl.pdf}
 	\vspace{-0.3cm}
 	\caption{ Workflow of self-supervised learning. 
 	%It automatically generates labels from input data and learns a representation network by fitting these auto-generated labels. The learned representation network is finetuned on the target task.
	}\label{fig:ssl}
	\end{center}
	\vspace{-0.3cm}
 \end{wrapfigure}
 \fi

Recently, in parallel to transfer learning, another pretraining paradigm -- self-supervised learning (SSL)~\cite{he2019momentum,chen2020simple} -- has arisen much research interest. Different from TL which is typically conducted in a supervised manner (by labels in source tasks), SSL is undertaken mostly in an unsupervised way without using any human-provided labels. The basic idea of SSL is to construct some auxiliary tasks solely based on the input data itself without using any human-offered annotations and encourage the network to learn meaningful representations by performing the auxiliary tasks well. Given the source data $\mathcal{D}_S$, SSL designs an auxiliary task $A$ and a network  $\mathcal{N}_A$ to perform this task.  $\mathcal{N}_A$ has a feature extraction sub-network $\mathcal{F}_A$ that has the same architecture as $\mathcal{F}_T$ and a prediction sub-network that is tailored to the task $A$. SSL trains $\mathcal{N}_A$ on $\mathcal{D}_S$ until convergence. Then it uses the weights of $\mathcal{F}_A$ to initialize $\mathcal{F}_T$, then trains $\mathcal{N}_T$ on $\mathcal{D}_T$ and $\mathcal{L}_T$ until convergence. Since SSL does not leverage labels in the source task, the learned feature extraction sub-network is not biased to these labels and is presumably more generalizable on $\mathcal{D}_T$. However, a potential downside is:  $\mathcal{F}_A$ is learned without human supervision, which renders the network less discriminative.

\vspace{-0.1cm}
\subsection{Contrastive self-supervised Learning}
\vspace{-0.2cm}
%\subsection{Self-supervised Learning}
%Self-supervised learning (SSL)~\cite{wu2018unsupervised,he2019moco,chen2020mocov2,chen2020simple} is a learning paradigm which aims to capture the intrinsic patterns and properties of input data (e.g., texts) without using human-provided labels. 
%The basic idea of SSL is to construct some auxiliary tasks solely based on the input data itself without using human-annotated labels and force the network to learn meaningful representations by performing the auxiliary tasks well. 
The auxiliary tasks in SSL can be constructed using many different mechanisms, such as rotation prediction~\cite{gidaris2018unsupervised},  image inpainting~\cite{pathak2016context},  automatic colorization~\cite{zhang2016colorful}, context prediction~\cite{nathan2018improvements}, etc. Recently, a contrastive mechanism~\cite{hadsell2006dimensionality} has gained increasing attention and demonstrated promising results in several studies~\cite{he2019momentum,chen2020simple}. The basic idea of contrastive SSL is: generate augmented examples of original data examples, create a predictive task where the goal is to predict whether two augmented examples are from the same original data example, and learn the representation network by solving this task. 
%to create pairs of data examples where the two examples in a pair are either labeled as being similar or dissimilar using prior knowledge, then train the network to correctly prediction whether two examples are similar or not. 

Different methods have been proposed to implement contrastive SSL. In SimCLR~\cite{chen2020simple} designed for image data, given the input images, random data augmentation is applied to these images. If two augmented images are created from the same original image, they are labeled as being similar; otherwise, they are labeled as dissimilar. Then SimCLR learns a network to fit these similar/dissimilar binary labels. The network consists of two modules: a feature extraction module $f(\cdot)$ which extracts the latent representation $\mathbf{h}=f(\mathbf{x})$ of an image $\mathbf{x}$  and a multi-layer perceptron $g(\cdot)$ which takes $\mathbf{h}$ as input and generates another latent representation $\mathbf{z}=g(\mathbf{h})$ used for predicting whether two images are similar. Given a similar pair $(\mathbf{x}_i,\mathbf{x}_j)$ and a set of images $\{\mathbf{x}_k\}$ that are dissimilar from $\mathbf{x}_i$, a contrastive loss can be defined as: $-\log(\textrm{exp}(\textrm{sim}(\mathbf{z}_i, \mathbf{z}_j)/\tau )/(\textrm{exp}(\textrm{sim}(\mathbf{z}_i, \mathbf{z}_j)/\tau )+\sum_{k}\textrm{exp}(\textrm{sim}(\mathbf{z}_i, \mathbf{z}_k)/\tau )))$, where $\textrm{sim}(\cdot,\cdot)$ denotes cosine similarity between two vectors and $\tau$ is a temperature parameter. SimCLR learns the network weights by minimizing losses of this kind. After training, the feature extraction sub-network is used for downstream tasks and $g(\cdot)$ is discarded. 

\iffalse
\begin{wrapfigure}{r}{0.42\textwidth}
    \centering
     %\vspace{-1.4cm}
  	\includegraphics[width = 0.42\columnwidth]{figs/moco.pdf}
    %\vspace{-0.3cm}
    \caption{Architecture of MoCo.}
    \label{fig:moco}
    %\vspace{-0.6cm}
\end{wrapfigure}
\fi

While SimCLR is easy to implement, it requires a large minibatch size to yield high performance, which is computationally prohibitive. MoCo~\cite{he2019momentum} addresses this problem by using a queue that is independent of minibatch size. This queue contains a dynamic set of  augmented data examples (called keys). In each iteration, the latest minibatch of examples are added into the queue; meanwhile, the oldest minibatch is removed from the queue. In this way, the queue is decoupled with minibatch size. 
%Figure~\ref{fig:moco} shows the architecture of MoCo. 
The keys are encoded using a momentum encoder. Given an augmented data example in the current minibatch (called query) and a key in the queue, they are considered as a positive pair if they originate from the same image, and a negative pair if otherwise.  A similarity score is calculated between the encoding of the query and the encoding of each key. Contrastive losses are defined on the similarity scores.

\vspace{-0.3cm}
\section{Design of Studies}
\vspace{-0.3cm}
In this section, we study several factors that may affect the comparative advantages between SSL and TL, including domain difference between source task and target task, the amount of pretraining data, class imbalance in source task, and usage of target data for additional pretraining. 
\vspace{-0.2cm}
\paragraph{Study of domain difference}
We use five source domains and four target domains, from the following areas: (1) objects in daily life, (2) scenes in daily life; (3) nature such as plants and animals, and (4) medical images such as CTs and X-rays. We use five source datasets:  ImageNet, SUN, iNaturalist, LUNA, and ChestX-ray8, from objects, scenes, nature, CT, and X-ray domains respectively. The details of these five datasets are:
\begin{itemize}[leftmargin=*]
    \item The ImageNet~\cite{deng2009imagenet} dataset contains 1,281,167 training and 50,000 validation images from 1,000 classes. These classes cover common objects in daily life, such as monitor, school bus, sleeping bag, teapot, broccoli, etc. 
    \item The SUN-397~\cite{sun} dataset contains 130,519 images from 397 scene categories, such as abbey, balcony, cafeteria, etc. Each category has more than 100 images. 
   % 899 categories and 130,519 images, with more than 100 images per category. 
%there are 397 semantic scene categories including abbey, diner, mosque, and stadium. It was introduced to provide a wide coverage of scene categories.
    \item The iNaturalist 2018 dataset~\cite{inaturalist} contains  437,513 training and 24,426 validation images from 8,142 fine-grained categories about nature, including plants, insects, birds, etc. The numbers of images in different categories are highly imbalanced. 
   \item The LUNA~\cite{LUNA} dataset contains 888 CT scans about lung nodules. 
  \item The ChestX-ray8~\cite{chestxray8} dataset contains 112,120 frontal-view X-ray images from 30,805 patients from 15 disease classes, such as emphysema, pneumonia, fibrosis, infiltration, cardiomegaly, etc.
\end{itemize}

We use four target datasets: Caltech-256, Flowers-102, COVID-CT, and Pneumonia from objects, nature, CT, and X-ray domains respectively. The detailed information of these datasets is as follows. 
%The three target domains are: general images, natural images, and medical images, represented by three datasets: Caltech-256, Flower-102, and COVID-CT.
%\subsubsection{Datasets}
%\begin{itemize}
   % \item The ImageNet~\cite{deng2009imagenet} contains 1,281,167 training and 50,000 validation images from 1,000 classes.
   % \item The SUN-397~\cite{sun} dataset contains 899 categories and 130,519 images, with more than 100 images per category. In the SUN-397 dataset, there are 397 semantic scene categories including abbey, diner, mosque, and stadium. It was introduced to provide a wide coverage of scene categories. 
  %\end{itemize}
\begin{itemize}[leftmargin=*]
    \item The Caltech-256~\cite{caltech} dataset contains 256 categories of objects in daily life, such as instruments, furniture, animals, food, vehicles, etc. The  number of training, validation, and testing images is 7710, 6425, and 6425 respectively. 
    \item The Flowers-102~\cite{flower} dataset contains 102 types of flowers. The  number of training, validation, and testing images is  6149, 1020, and 1020 respectively. 
    \item The COVID-CT~\cite{zhao2020covid} dataset contains  349 COVID-19 CT images from 216 patients and 463 non-COVID-19 CTs. The  number of training, validation, and testing images is 425, 118, and 203.
    \item  The Pneumonia~\cite{kermany2018identifying} dataset contains 5,863 X-ray images which are either positive for pneumonia or negative. The  number of training, validation, and testing images are 5216, 16, and 624 respectively.
    % Pneumonia Chest X-ray dataset were selected from retrospective cohorts of pediatric patients of one to five years old from Guangzhou Women and Children’s Medical Center, Guangzhou. There are 5,863 X-Ray images and 2 categories (Pneumonia/Normal).
\end{itemize}

%\begin{itemize}
 %   \item ++: visual distance < 1.6 and  label similarity > 0.6
%\end{itemize}

We perform a quantitative measurement of domain difference of two tasks on their visual contents and class labels. To measure visual distance between domains, we use the  method in~\cite{domain_difference}. We sample 1000 images from a source task and label them as 0, and sample 1000 images from a target task and label them as 1. Then we split the 2000 images into a training set and a test set. We train a classifier on the training set to distinguish whether an image is from source or target, then measure the classification error $\epsilon$ on the test set. The visual  difference is defined as $d=2(1-2 \epsilon)$. Intuitively, if source images and target images are easy to be told apart (i.e., $\epsilon$ is small), then their visual difference is large. 
 %As can be seen, these scores are consistent with human's judgement about domain difference. 
In addition to visual difference, we also measure domain difference in the label space. Given $\{(c_i,f_i)\}_{i=1}^m$ in the source dataset, where $c_i$ is a class name, $f_i$ is the frequency of this class in this dataset, and $m$ is the number of classes in the source dataset, and similarly  $\{(c_j,f_j)\}_{j=1}^n$ in the target dataset, we calculate the class similarity of these two domains using this equation: $(\sum_{i=1}^m\sum_{j=1}^n f_if_j \sigma(c_i,c_j))/(\sum_{i=1}^m f_i \sum_{j=1}^n f_j)$, where $\sigma(c_i,c_j)$ is the cosine similarity between the GloVe~\cite{pennington2014glove} embeddings of $c_i$ and $c_j$. This equation basically measures the average similarity of the labels of a source image and a target image.

\begin{wrapfigure}{r}{0.75\textwidth}
    \centering
    \small
    % \vspace{-0.2cm}
    \begin{tabular}{lcccc}
    \toprule
         %\diagbox[width=8em]{Source}{Target} 
         & Caltech-256 &Flowers-102 & COVID-CT & Pneumonia
         \\ \midrule
          \textbf{Visual distance} & & & & \\
         ImageNet &1.35 &1.73 & 1.69&1.89
         \\ 
         SUN &1.09  &1.67 &1.71 & 1.84
         \\ 
         iNaturalist & 1.34& 1.32& 1.75 &  1.89
         \\ 
         LUNA & 1.81 &1.90 & 1.30& 1.90
         \\
         ChestX-ray8 & 1.99 &1.83 & 1.81 & 1.54\\
         \hline
         \hline
          \textbf{Class similarity} & & & & \\
          ImageNet &0.07 & 0.06&0.02 & 0.01
         \\ 
         SUN &0.07  & 0.04&0.04 & 0.02
         \\ 
         iNaturalist &0.03 &0.07 & -0.11 & -0.07 
         \\ 
         LUNA & 0.05&0.01 & 0.62& 0.57
         \\
         ChestX-ray8 & 0.01 &0.01 & 0.23 & 0.25
         \\
         \hline
         \hline
           ImageNet & + + & - +  & - - & - - 
         \\ 
         SUN & + + & - -  & - - & - -
         \\ 
         iNaturalist &   + - & + + & - - & - - 
         \\ 
         LUNA & - -  & - - & + + & - +
         \\
         ChestX-ray8 & - - & - - & - +  & + +
         \\ \bottomrule
        %  Rand & \multicolumn{2}{c|}{50.75} & \multicolumn{2}{c|}{59.2} & \multicolumn{2}{c}{61.41}
        %  \\ \hline
    \end{tabular}
        \caption{Source-target domain difference. }
    \label{tab:difference measure}
    \vspace{-0.5cm}
\end{wrapfigure}

Figure~\ref{tab:difference measure} shows the visual difference and class similarity between source and target tasks. If the visual distance is less than 1.6, a source and a target are considered as in the same domain (denoted by "+"); otherwise, in different domains (denoted by "-"). For class similarity, "+" if the similarity is greater than 0.6, and "-" if otherwise.
%If the class similarity is greater than 0.6,
%If the class similarity is greater than 0.6, two tasks are in the same domain (denoted by "+"); otherwise, in different domain (denoted by "-"). 
The third panel in Figure~\ref{tab:difference measure} shows the results, where the first and second symbol in each cell correspond to visual distance and class similarity respectively. A source and a target with "++" have high domain similarity, and those with "- -" have large domain differences. The results are in accordance with intuitive judgement. For example, ImageNet and Caltech-256 are in the same domain about daily objects; iNaturalist and Flowers-102 are in the same domain about nature. 
%Between the source and target tasks, ImageNet and Caltech-256 are in the same domain about daily objects; iNaturalist and Flowers-102 are in the same domain about nature; LUNA and COVID-CT are in the same domain about medicine. 

%We construct a new dataset $U=\left\{\left(\mathbf{x}_{i}, 0\right)\right\}_{i=1}^{1000} \cup\left\{\left(\mathbf{x}_{i}, 1\right)\right\}_{i=1001}^{2000}$, where $\left\{\mathbf{x}_{i} \right\}_{i=1}^{1000}$ are examples of the source sample labeled as 0 and $\left\{\mathbf{x}_{i} \right\}_{i=1001}^{2000}$ are examples of the target sample labeled as 1. We randomly partition the images into a training set and testing set with a ratio of 0.7 and 0.3, and we train a simple classifier with the generalization error of $\epsilon$ on discriminating between source and target examples. We use the Proxy A-distance (PAD) given by $d=2(1-2 \epsilon)$ to quantify the difference. Table~\ref{tab:difference measure} shows the domain difference between source and target tasks. 

\vspace{-0.3cm}
\paragraph{Study of the amount of pretraining data}
To study how SSL and TL are affected by the amount of pretraining data, we perform the following controlled study: for each source dataset, we use 1\%, 10\%, and 100\% of the dataset for pretraining. 

\vspace{-0.3cm}
\paragraph{Study of class imbalance in source tasks} In a source task with class labels, the frequencies of classes are typically imbalanced. We are interested in studying how class imbalance affects the performance of SSL and TL. From a source dataset, we create imbalanced datasets with different imbalance ratios, pretrain SSL and TL on these datasets, and check how their performance varies with the imbalance ratio. Given a source dataset containing $K$ classes, we rank these classes in ascending order of their frequencies. Let $n_j$ denote the frequency of the $j$-th ranked class. The class imbalance ratio $\rho$ is defined as $n_{K}/n_{1}$, which is the ratio between the number of training examples of the most and least frequent class. We create datasets with different imbalance ratios by tuning $n_{K}$ and $n_1$.  For the rest  classes, we perform random sampling to ensure the frequencies of these classes form an arithmetic sequence: specifically, the frequency of class $j$ is $(j-1)\times(n_K-n_1)/(K-1)+n_1$, for $j=2,\cdots,K-1$.

%We investigate how the degree of class imbalance in source tasks affect the performance of TL and SSL. We also investigate how the source task data imbalance influences the performance on the target task. We construct 3 imbalanced subsets of ImageNet with a ratio of $\rho = \{1,5,25\}$. We define the imbalance ratio as $\rho=\max _{i}\left\{n_{i}\right\} / \min _{i}\left\{n_{i}\right\}$, which is the ratio between sample number of the most frequent and least frequent class.
\vspace{-0.3cm}
\paragraph{Study of using target data for additional pretraining} In existing approaches, pretraining is mostly conducted on source data. We are interested in investigating whether it is helpful  to add the training examples in the target task as additional data for pretraining and how this will affect the comparative advantages of SSL and TL. We compare the following pretraining settings: (1) SSL on source data only, on target data only, and on the combined data of source and target; (2) TL on source data only, on target data only, and on the combined data of source and target. For SSL on target data only, we pretrain SSL on training images in the target task, then finetune using both input images and output labels in the target training set. For SSL on the combined data of source and target, we pretrain SSL on source images and target training images. For TL on target data only, it is the same as finetuning on the target task with random initialization of the network. For TL on the combined data, it is a multi-task learning problem which trains classification models for the source task and target task simultaneously.

\iffalse

1. Source and target tasks have different class labels and large domain shift \\
Target task: COVID-CT classification\\
Source task: ImageNet classification\\
Compare TL and SSL\\

2. Source and target tasks have different class labels and small domain shift \\
Target task: COVID-CT classification\\
Source task: LUNA classification\\
Compare TL and SSL\\

3. Source and target tasks have the same class labels and large domain shift\\
Target task: CT-based lung nodule classification\\
Source task: chest x-ray based lung nodule classification\\
Compare TL and SSL\\

4. Source and target tasks have the same class labels and small domain shift\\
Target task: NIH chest X-ray pneumonia classification\\
Source task: Stanford chest X-ray pneumonia classification\\
Compare TL and SSL\\

Combine TL and SSL:

How class imbalance affects the performance of TL and SSL:
\fi

\vspace{-0.2cm}
\section{Experiments}
\vspace{-0.2cm}
%We perform SSL and TL pretraining on different source datasets and on different amount of source data. Then we finetune these pretrained models on target tasks and report performance on test data in target tasks. 
%We evaluate the performance of transfer learning by using ImageNet, Sun, iNat, and LUNA as source domains, and Caltech-256, Flowers-102 and COVID-CT as target domains. 

\subsection{Experimental settings}
\vspace{-0.2cm}
All source tasks and target tasks are about image classification. We used ResNet-50~\cite{resnet} for classification. The architecture of ResNet-50 is the same for different tasks, except the number of output units (which is the same as the number of categories in a task). The  TL and SSL models pretrained on the full ImageNet dataset were borrowed from \cite{resnet} and \cite{he2019momentum}. 
%For TL and SSL comparison on quantity difference, we use the pretrained model provided in \cite{resnet} and \cite{MoCo} for $100 \%$ ImageNet experiment. 
The rest of the models were pretrained from scratch. 

For TL on source tasks, we applied data augmentation including random resize with a ratio sampled from $[0.08,1]$, random crop of $224\times 224$, and the ImageNet AutoAugment policy~\cite{cubuk2019autoaugment}. The weights were optimized using SGD with an initial learning rate of $0.1$ and a momentum of $0.9$. We pretrained TL for 200 epochs on every source task with a minibatch size of 128. The learning rate was reduced by 0.1 every 50 epochs. 

For SSL on source tasks, we used the data augmentation methods in~\cite{chen2020mocov2} and used SGD as the optimizer with an initial learning rate of $0.015$ and a momentum of $0.9$. We pretrained SSL for 200 epochs and a batch size of 128. The learning rate was adjusted using cosine learning rate scheduling. 

For finetuning on target tasks, we utilized the same data augmentation methods as in TL pretraining. We used Adam~\cite{adam} as the optimizer with an  initial learning rate of 0.001. The finetuning was performed for 200 epochs on Caltech-256 and Flower-101, 100 epochs on COVID-CT, and 30 epochs on Pneumonia. Cosine  scheduling with a period of 6 was used to adjust the learning rate. The implementation was based on PyTorch and the experiments were conducted on 8 Tesla V100 GPUs.

\vspace{-0.1cm}
\subsection{Results on domain difference}
\vspace{-0.2cm}
We perform controlled studies to investigate how domain difference between source and target tasks affects the performance of SSL and TL. To rule out the influence of the amount of pretraining data, we make the number of pretraining examples in all source tasks the same. This number is chosen to be 90,000. For a source dataset whose size is larger than 90,000, we randomly sample 90,000 data examples from this dataset.

%We evaluate the effect of domain difference by pre-training the network on different source datasets from scratch and then finetune on different target dataset with both SSL and TL.The experiment results are shown in Table~\ref{tab:different-ssl}.
\begin{table}[H]
    \centering
     \caption{Results on domain difference. Rows correspond to source datasets where SSL and TL are pretrained. Columns correspond to target datasets where pretrained models by SSL and TL are finetuned. The numbers are top-1 image classification accuracy achieved on the test data in target tasks. The number of pretraining data examples in different source datasets is the same.}
    \begin{tabular}{lcccccccc}
    \toprule
         %\diagbox[width=8em]{Source}{Target} 
         & \multicolumn{2}{c}{Caltech-256} &\multicolumn{2}{c}{Flowers-102} & \multicolumn{2}{c}{COVID-CT} & \multicolumn{2}{c}{Pneumonia}
         \\ \cmidrule(r){2-3} \cmidrule(r){4-5} \cmidrule(r){6-7} \cmidrule(r){8-9} 
         & SSL & TL & SSL & TL & SSL & TL & SSL & TL
         \\ \midrule
         ImageNet & 57.53 & 62.14& 82.02 &86.23 & 76.01 &73.08 & 94.07 & 92.47
         \\ 
         SUN & 56.34 &62.21 & 79.55 & 63.34& 80.33 & 71.24& 93.75& 92.41
         \\ 
         iNaturalist & 56.59 & 47.53 & 82.19 &86.47 & 79.59 &74.48 & 94.87 & 93.59
         \\ 
         LUNA & 53.43 & 48.31 & 76.77 & 51.40 & 79.11 & 74.04 & 93.43 & 93.27
         \\
         ChestX-ray8 & 53.50 & 47.31 & 72.09 & 55.40 & 78.63 & 72.20 & 93.27 & 95.19
         \\ \bottomrule
        %  Rand & \multicolumn{2}{c|}{50.75} & \multicolumn{2}{c|}{59.2} & \multicolumn{2}{c}{61.41}
        %  \\ \hline
    \end{tabular}
    \label{tab:Transfer}
    \vspace{-0.2cm}
\end{table}

%Each row represents a network pre-trained on a specific source domain, and each column shows the \textbf{top-1} image classification accuracy by fine-tuning different networks on a target fine-grained dataset.

Table~\ref{tab:Transfer} shows the results on domain difference.  From this table, we make the following observations. On most source-target pairs with strong domain similarity ("++"), including (ImageNet, Caltech-256), (SUN, Caltech-256), (iNaturalist, Flowers-102), (ChestX-ray8, Pneumonia), TL achieves better accuracy than SSL. This shows that when the domain difference is small, TL is more effective than SSL. The possible reason is: when the domain difference is small, the classes in source labels have a lot of overlap with those in target labels. Due to this overlap, it is beneficial to transfer the semantic information in  source labels to the target task. TL utilizes sources labels while SSL does not. Hence TL works better than SSL. The only exception is (LUNA, COVID-CT), where TL performs worse than SSL. Second, on source-target pairs where the domain difference is large ("- -"), including (ImageNet, COVID-CT), (ImageNet, Pneumonia), (SUN, Flowers-102), (SUN, COVID-CT), (SUN, Pneumonia), (iNaturalist, COVID-CT), (iNaturalist, Pneumonia), (LUNA, Caltech-256), (LUNA, Flowers-102), (ChestX-ray8, Caltech-256), (ChestX-ray8, Flowers-102), SSL achieves better accuracy than TL. This shows that when domain difference is large, SSL works better than TL. The possible reason is: when the domain difference is large, the classes in source labels are largely different from those in target labels. When the representations are learned by fitting source labels, they are biased to source classes and generalize less well on the target task which has a different set of classes. SSL avoids using source labels, hence is not prone to such a bias.  Third, SSL is less sensitive to domain difference than TL. For example, on Caltech-256, the performance of SSL is relatively stable. The difference between the highest and lowest performance is about 4\% (absolute). In contrast, for TL, the difference between the highest and lowest performance is about 15\% (absolute). This is because TL uses source labels for pretraining. When the domain difference varies substantially, the labels change substantially, which makes the learned representations vary a lot. Fourth, on the same target domain, the closer a source domain is to this target, the better the performance is, for both SSL and TL. For example, in terms of domain similarity with Caltech-256, we have the following order: ImageNet, SUN $>$ iNaturalist $>$ LUNA, ChestX-ray8. As can be seen, the accuracy corresponding to these source domains decreases in general. This is because pretraining on a closer source domain can make the learned representations more suitable for the target domain.

\vspace{-0.2cm}
\subsubsection{Results on the amount of pretraining data}
\vspace{-0.2cm}
We evaluate the effect of the amount of pretraining data  by
pretraining SSL and TL on different subsets of each source dataset. Given a source dataset, we create two subsets by randomly sampling 1\% and 10\% of data examples.  The experimental results are shown in Table~\ref{tab:amount}.
%the network on the same source dataset with different partition of the source dataset --- 1\%, 10\% and 100\%, from scratch and then finetune on different target datasets via both SSL and TL strategy. The experiment results are shown in Table~\ref{tab:Transfer}.
From this table, we make the following observations. First, when the amount of training data is small, SSL outperforms TL. For example, pretrained on 1\% ImageNet, SSL outperforms TL on all target tasks. One possible reason is: when the training data is small, TL has a risk of overfitting to the labels of the small dataset and generalizes less well on target tasks. In contrast, SSL is an unsupervised pretraining method, which does not have the risk of overfitting to labels of small-sized source data. Second, when the amount of pretraining data is large, TL outperforms SSL. For example, pretrained on 100\% ImageNet data, TL achieves better performance than SSL on all target tasks. The possible reason is: when the dataset is large, TL is less likely to suffer overfitting. On the contrary, the diverse labels contained in the large dataset enables TL to learn discriminative representations. 
%Third, when domain difference is large, SSL outperforms TL. For example, pretrained on 100\% SUN, SSL achieves much better performance than TL on COVID-CT. 
Third, when domain difference is small, SSL is less sensitive to data amount than TL. For example, when the pretraining ImageNet data increases from 1\% to 100\%, the relative improvement of SSL on Caltech-256 is about 30\% whereas the relative improvement of TL is about 62\%. Fourth, for both SSL and TL, increasing the amount of training data leads to better performance. This is not surprising since deep learning methods are data hungry. Fifth, when the pretraining data is small, the performance of SSL and TL may be worse than random initialization. This is because SSL and TL may be overfitted to the small-sized pretraining data and generalize  less well on the target tasks.   

\begin{table}[H]
    \centering
    \caption{Results on the amount of pretraining data. Rows correspond to datasets for pretraining. Columns correspond to target datasets for finetuning.}
    % \begin{tabular}{{2.5cm}{1cm}{1cm}{1cm}{1cm}{1cm}{1cm}}
    \small
        \begin{tabular}{llccccccccc}
    \toprule
         & & & \multicolumn{2}{c}{Caltech-256} &\multicolumn{2}{c}{Flowers-102} & \multicolumn{2}{c}{COVID-CT}& \multicolumn{2}{c}{Pneumonia}
         \\ \cmidrule(r){4-5} \cmidrule(r){6-7} \cmidrule(r){8-9} \cmidrule(r){10-11} 
         & & Number & SSL & TL & SSL & TL & SSL & TL& SSL & TL 
         \\ \midrule
         1\%  & ImageNet & 10,000 & 52.33&49.87 & 73.13&63.05 &75.26 &66.39 & 93.59 & 90.54
          \\ 
          10\% &  ImageNet & 100,000 &57.78 &69.09 &81.27 & 87.09&78.85 &72.86& 94.55 & 93.43
          \\ 
         100\% & ImageNet & 1,000,000 &68.16& 80.62 & 87.52&91.48 &80.00 &82.03& 94.55 & 95.67
         \\ 
         1\% & SUN &1,088 & 26.73&31.74 & 38.84& 55.87&55.11 &67.65& 87.50 & 86.21
         \\ 
         10\% & SUN &10,875 & 40.29 & 52.61& 56.46&64.88 & 71.95&68.18& 89.74 & 86.54
         \\ 
         100\% & SUN & 108,754& 53.92 &60.46 &78.12 & 83.75&79.37 &73.61& 93.59 & 91.19
         \\ 
         1\% & iNaturalist & 4,375 & 52.85 &38.97 & 75.99 & 64.50& 74.56 &69.54& 90.97 & 89.10
         \\ 
         10\% & iNaturalist & 40,710& 54.18 & 45.83 &76.38 &78.46 & 79.37 &71.68& 91.19 & 92.47
         \\ 
         100\% & iNaturalist & 437,513 &60.48 & 60.23 & 86.33 &88.60 & 80.77 &74.30& 95.35 & 93.75
         \\ 
         1\% & LUNA & 895 & 43.94 & 46.38 & 55.50 & 52.85 & 65.90 & 66.17& 87.98 & 86.70
         \\ 
         10\% & LUNA & 8,945 & 47.25 & 44.79 & 64.01 & 52.90 & 72.86 & 71.90& 88.62 & 87.34
         \\ 
         100\% & LUNA & 89,455 & 53.43 & 48.31 & 76.77 & 54.44 & 79.11 & 74.04 & 93.43 & 93.27
         \\ 
         1\% & ChestX-ray8 & 900 & 51.29 & 31.93 & 45.88 & 50.86 & 57.82 & 64.99 & 91.99 & 90.87
         \\ 
         10\% & ChestX-ray8 & 9,000 & 50.64 & 50.36 & 56.80 & 54.77 & 70.45 & 67.79 & 92.79 & 93.27
         \\ 
         100\% & ChestX-ray8 & 89,953 & 53.50 & 47.31 & 72.09  & 57.81 & 78.63 & 72.20 & 93.27 & 95.19
         \\ \midrule
        \multicolumn{2}{l}{Random initialization}  & &
         \multicolumn{2}{c}{50.75} & \multicolumn{2}{c}{59.26} & \multicolumn{2}{c}{65.51} &
         \multicolumn{2}{c}{88.62}
         \\ \hline
    \end{tabular}
    \label{tab:amount}
    \vspace{-0.4cm}
\end{table}

%\subsection{Data Quantity Discrepancy}

%This may due to differences of feature representations between the combined data and source data, and features extracted from the source data are more appropriate for source data classification than features extracted from the combined data. 

\vspace{-0.2cm}
\subsection{Results on class imbalance in source tasks}

\begin{wrapfigure}{r}{0.75\textwidth}
% \begin{wrapfigure}{r}{0.75\textwidth}
    \centering
    \vspace{-0.4cm}
       \small
   \begin{tabular}{ccccccccc}
    \toprule
    &\multicolumn{2}{c}{Caltech-256} &\multicolumn{2}{c}{Flowers-102} & \multicolumn{2}{c}{COVID-CT}& \multicolumn{2}{c}{Pneumonia}
         %\diagbox[width=8em]{Source}{Target} 
         \\ \cmidrule(r){2-3} \cmidrule(r){4-5} \cmidrule(r){6-7}  \cmidrule(r){8-9} 
      Ratio    & SSL & TL & SSL & TL & SSL & TL & SSL & TL
         \\ \midrule
         1 &59.34 &66.62 &84.45 &88.06 & 80.29 &79.33 & 95.35 &94.43
          \\ 
         5 & 58.72  &65.77  &83.12  &86.72  & 79.81  &74.04  &94.87 &93.91 
          \\ 
         25 & 58.46  &64.88 & 82.68 &86.10 &75.96  &72.86 & 94.71  & 93.75 
        %  \\
        %  \midrule
        %  1 & - & - & - &- & - &- & - & -
        %   \\ 
        %  5 & -0.62 &-0.85 & -1.33 &-1.34 & -0.48 &-5.29 &-0.48&-0.52
        %   \\ 
        %  25 & -0.88&-1.74 & -1.77 &-1.94 &-4.33 &-6.47& -0.64 & -0.67
         \\ \bottomrule
    \end{tabular}
        \caption{Comparison of results under different imbalance ratio }
    \label{tab:imbalance}
    \vspace{-0.8cm}
% \end{wrapfigure}
\end{wrapfigure}

 From ImageNet, we create 3 imbalanced datasets, with imbalance ratio $\rho=1, 5, 25$, respectively. For $\rho=1$, we set $n_1=n_{N}=130$; for $\rho=5$, $n_1=43$, $n_{N}=215$; for $\rho=25$, $n_1=10$, $n_{N}=250$. Such settings ensure the total number of data examples in each dataset to be approximately the same ($\approx 13,000$), so that the factor of pretraining data size can be ruled out.

Figure~\ref{tab:imbalance} shows the results under different class imbalance ratios. From this table, we make the following observations. First, as the imbalance ratio increases, the performance of both SSL and TL decreases. The reason is that class imbalance renders the learned representations biased to images from frequent classes in the source task. The more imbalanced the classes are, the larger the bias is. A larger bias leads to worse generalization to target tasks. Second, SSL is more robust to class imbalance than TL. For example, on the COVID-CT task, the gap between the best and worst performance of SSL under different imbalance ratios is about 4.3\% (absolute) while the gap for TL is about 6.5\% (absolute). Similar results are observed on other target tasks as well. The reason is: TL is pretrained using  class labels and therefore is more sensitive to the distribution of labels, including imbalance ratio. In contrast, the pretraining of SSL is label-free, hence is less affected by label distribution. 

\vspace{-0.2cm}
\subsection{Results on using target data for pretraining}
\vspace{-0.2cm}

%To investigate the effect of different self-supervised pre-training under different pre-training data configuration, we perform self-supervised pre-training on (1) Source dataset (2) Target dataset (3) Combined datase and finetuning on the target task. We compare the target task performance to see which self-supervised strategy can most effectively improve representation ability of the model. 
\begin{wrapfigure}{r}{0.75\textwidth}
    \centering
    \vspace{-0.3cm}
    \small
    \begin{tabular}{cccccccc}
    \toprule

         \multirow{2}{*}{Source}&  \multirow{2}{*}{Target} & \multicolumn{2}{c}{Source only} &  \multicolumn{2}{c}{Target only} &
         \multicolumn{2}{c}{Combined} 
         \\ \cmidrule(r){3-4} \cmidrule(r){5-6} \cmidrule(r){7-8} 
         & &SSL & TL &   SSL &TL   & SSL  &  TL 
         \\ \midrule
         SUN & Caltech-256 &53.92&60.46 &56.59 &50.75& 57.46 &59.57
         \\
         SUN & Flowers-102 & 78.12&83.75&57.59 &59.26&79.69 &78.00\\ 
         SUN & COVID-CT & 79.37&73.61& 61.72 &65.51& 80.77 &69.74
         \\
         SUN & Pneumonia & 93.75&91.19 & 93.43 &88.62& 94.39&90.04
         \\ \bottomrule
    \end{tabular}
    \caption{Results on the usage of target data for additional pretraining.}
    \vspace{-0.2cm}
    \label{tab:different-ssl}
\end{wrapfigure}
In this study, we compare the following pretraining settings: on source data only, on target data only, and on the combined data of source and target. The source dataset is SUN and target data refers to the data examples in the training set of a target task. Figure~\ref{tab:different-ssl} shows the results. From this table, we make the following observations. First, for SSL, pretraining on the combined data performs better than on source-only and on target-only. The possible reason is: combined data has more training examples, which helps to learn better representations. Second, TL on the combined data performs better than on target-only. The reason is that performing TL on the combined data effectively leverages the source task to help with the learning of the target task via multi-task learning. Third, TL on the combined data performs worse than on source-only. The reason is: TL on source-only first pretrains on source data, then finetunes on target data. The focus is finetuning on target tasks, where the network is learned to best fit the target data. For TL on the combined data, training on source task and target task is performed simultaneously. The network aims to perform both tasks well instead of focusing on the target task. The reduced focus incurs worse performance on target tasks.

\vspace{-0.2cm}
\section{Related Works}
\vspace{-0.3cm}
Several works conducted preliminary comparisons of SSL and TL. In \cite{he2019momentum}, supervised transfer learning (TL) is compared with an SSL approach -- MoCo. It is shown that in certain tasks SSL outperforms TL, but the other way around in other tasks. It is not studied what factors incur such inconsistent results, which renders ML researchers have to try both approaches to find out which one works better empirically without an informed guidance. We aim to bridge this gap in our work, by systematically investigating under what situations SSL performs better than TL and vice versa. Zhai et al.~\cite{zhai2019visual} compared a number of representation learning methods, including supervised TL and SSL. They compared SSL and TL in terms of a single factor: image style. In our work, we perform a comprehensive comparison of SSL and TL on a number of factors. Newell and Deng~\cite{newell2020useful} studied what factors affect the performance of SSL. The studies are performed on SSL only, rather than on the comparison with TL.  Resnick et al.~\cite{resnick2019probing} performed a detailed study of SSL and presented several findings such as linear probes is insufficient to evaluate the representations learned by SSL, the bias towards success stemming from the architecture is high, etc. The study is mostly focused on SSL, instead of comparing with TL.  Our work differs from these two in that our goal is to study what factors affect the comparative advantages between TL and SSL.

\vspace{-0.2cm}
\section{Conclusions}
\vspace{-0.3cm}
In this work, we make a comprehensive comparative study about SSL and TL, regarding which method works better under different properties of data and tasks. Specifically, we study how the domain difference between source and target tasks, the amount of pretraining data, class imbalance in source tasks, and the usage of target data for additional pretraining affect the comparative advantages of SSL and TL. On 5 source tasks and 4 target tasks from various domains, on  varying amount of pretraining data, we conduct experiments and distill a set of insights. These insights can potentially help ML researchers to decide which pretraining method to use based on the properties of their applications and foster the development of new SSL and TL methods.

% \section*{Broader Impact}
% Our work studies how to learn better representations of images. It can potentially benefit many communities in the society. For example, hospitals can leverage methods investigated in our study to analyze medical images for AI-aided diagnosis of diseases. As another example, these methods can be used to detect violence and pornography photos on social media. On the other hand, these methods may be misused by malicious users for performing adversarial attacks on self-driving systems, creating fake photos, etc. Such misuse should be closely monitored and prevented timely. If these methods fail to learn good representations of images, they may lead to incorrect predictions. In mission critical applications, the usage of these methods should be guided by experienced ML experts and domain experts. 

\bibliographystyle{unsrt}
\bibliography{refs}

\end{document}